\documentclass[letterpaper, 10 pt, journal, twoside]{ieeetran}

\usepackage[T1]{fontenc}
\usepackage{amsmath,amsfonts}
\usepackage{algorithm}
\usepackage{array}

\newcolumntype{L}[1]{>{\raggedright\let\newline\\\arraybackslash\hspace{0pt}}m{#1}}
\newcolumntype{C}[1]{>{\centering\let\newline\\\arraybackslash\hspace{0pt}}m{#1}}
\newcolumntype{R}[1]{>{\raggedleft\let\newline\\\arraybackslash\hspace{0pt}}m{#1}}

\usepackage[caption=false,font=normalsize,labelfont=sf,textfont=sf]{subfig}
\usepackage{textcomp}
\usepackage{url}
\usepackage{verbatim}
\usepackage{graphicx}
\usepackage{cite}

\usepackage{amssymb}  
\usepackage{xspace}
\usepackage{booktabs}
\usepackage{xcolor}
\usepackage{flushend}
\usepackage[font=small]{caption}
\usepackage{multirow}
\usepackage{etoolbox}
\usepackage{hyperref}
\usepackage{multirow}

\begin{document}

\newcommand{\highlight}[1]{{{#1}}}

\newcommand{\papername}{SeMLaPS}
\newcommand{\methodname}{\papername\xspace}

\makeatletter
\apptocmd{\@maketitle}{\centering \includegraphics[width=0.98\linewidth]{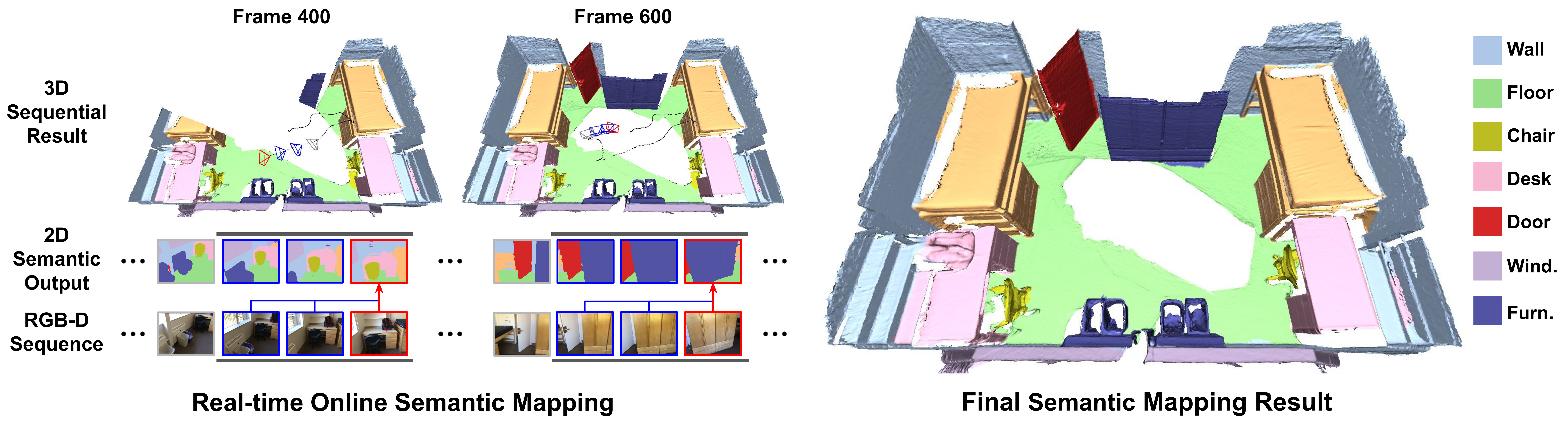}
\vspace{-0.2em}
\captionof{figure}{\methodname takes an RGB-D sequence as input and incrementally builds a semantic map of the scene. Our Latent Prior Network (LPN) outputs 2D semantic labels that encode history information of prior views via differentiable feature re-projection. 2D labels are lifted to 3D and fused temporarily via Bayesian fusion. Based on a novel geometric map over-segmentation, our Segment-Convolutional Network further refines the 3D semantic map by applying convolutions at the  segment-level, resulting in state-of-the-art accuracy within real-time 2D-3D networks, and good cross-sensor generalization capabilities at the same time. \vspace{-6mm} \label{fig:teaser}}
}{}{}
\makeatother

\title{\papername: Real-time Semantic Mapping with Latent Prior Networks and Quasi-Planar Segmentation}

\author{Jingwen Wang$^{1,2}$, Juan Tarrio$^{1}$, Lourdes Agapito$^{2}$, Pablo F. Alcantarilla$^{1}$, 
Alexander Vakhitov$^{1}$

\thanks{Manuscript received June 13, 2023; revised August 25, 2023; Accepted September 28, 2023.This paper was recommended for publication by Editor Markus Vincze upon evaluation of the Associate Editor and Reviewers’ comments.}%
\thanks{$^{1}$ Slamcore Ltd. $^{2}$ University College London}
\thanks{Digital Object Identifier (DOI): see top of this page.}}

\markboth{IEEE Robotics and Automation Letters. Preprint Version. Accepted September, 2023}
{Wang \MakeLowercase{\textit{et al.}}: \papername: Real-time Semantic Mapping with Latent Prior Networks and Quasi-Planar Segmentation} 


\maketitle

\begin{abstract}
The availability of real-time semantics greatly improves the core geometric functionality of SLAM systems, enabling numerous robotic and AR/VR applications. We present a new methodology for real-time semantic mapping from RGB-D sequences that combines a 2D neural network and a 3D network based on a SLAM system with 3D occupancy mapping. When segmenting a new frame we perform latent feature re-projection from previous frames based on differentiable rendering. Fusing re-projected feature maps from previous frames with current-frame features greatly improves image segmentation quality, compared to a baseline that processes images independently. For 3D map processing, we propose a novel geometric quasi-planar over-segmentation method that groups 3D map elements likely to belong to the same semantic classes, relying on surface normals. We also describe a novel neural network design for lightweight semantic map post-processing. Our system achieves state-of-the-art semantic mapping quality within 2D-3D networks-based systems and matches the performance of 3D convolutional networks on three real indoor datasets, while working in real-time. Moreover, it shows better cross-sensor generalization abilities compared to 3D CNNs, enabling training and inference with different depth sensors. Code and data can be found at \url{https://github.com/slamcore/semlaps}. 
\end{abstract}


\setcounter{figure}{1}
\begin{figure*}
\vspace{1.0em}
\includegraphics[width=\textwidth]{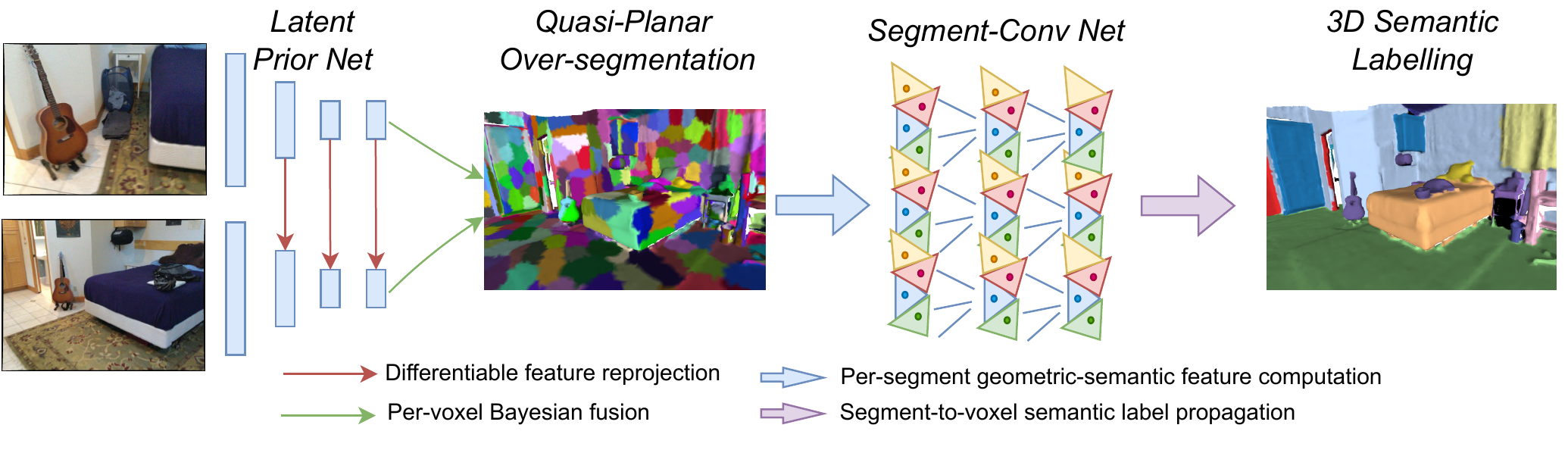}
\vspace{-1.75em}
\caption{\methodname runs on top of a state-of-the-art visual inertial SLAM system performing camera pose tracking and voxel-based 3D geometry reconstruction. \methodname has three main stages: First, a Latent Prior Network (LPN), relying on a fast 2D neural network backbone enhanced with differentiable latent feature re-projection, processes the RGB-D frames. It outputs 2D semantic labels, which are further lifted to 3D and aggregated over time with Bayesian fusion~\cite{mccormac2017semanticfusion}. Next, Quasi-Planar Over-Segmentation (QPOS) groups surface voxels into segments, associated with semantic class probabilities and geometric features. Finally, our Segment-Convolutional Network (SegConvNet) applies convolutions at segment-level and obtains final semantic labels. \vspace{-1.5em}}\label{fig:intro}
\end{figure*}
\vspace{-1.0em}
\section{Introduction}

\IEEEPARstart{S}{LAM} systems use semantics for better pose estimation or re-localization~\cite{ren2022visual,tian2022kimera} or to work in dynamic scenes~\cite{ren2022visual,tian2022kimera}. It can facilitate downstream tasks such as robotic navigation~\cite{kantaros2022perception}
or augmented reality (AR) experiences~\cite{narita2019panopticfusion}. 

Real-time semantic mapping methods usually rely on 2D convolutional neural networks with optional 3D post-processing (2D-3D networks) to annotate incoming images with semantics, using back-projection to lift the semantic labels to the 3D map~\cite{huang2021supervoxel,tian2022kimera,zhang2020fusion,narita2019panopticfusion,grinvald2019volumetric,mccormac2017semanticfusion},
while recent FP-Conv~\cite{zhang2020fusion} or SVCNN~\cite{huang2021supervoxel} also rely on lightweight 3D post-processing. 2D-3D networks  repetitively process images with similar visual content, solving 2D semantic segmentation from scratch for each image, which may be redundant~\cite{xiang2017darnn};  lack multi-view consistency in 2D labels~\cite{ma2017multi}; suffer from occlusions or object scale uncertainty~\cite{han2020occuseg}. Direct processing of 3D geometry with 3D convolutions, e.g. with MinkowskiNet~\cite{choy20194d} or SparseConvNet~\cite{graham20183d}, outperforms state-of-the-art 2D-3D networks. However, many semantic SLAM tasks require 2D image-level semantic labels, e.g. for semantic landmark annotation as in~\cite{ren2022visual} or for semantics-based processing of dynamics as in~\cite{bescos2021dynaslam}. This makes the use of 3D networks as main semantic inference engines in SLAM systems less appealing, even given recent progress towards using 3D networks in an online and incremental fashion~\cite{liu2022ins}. 

Semantic SLAM systems may struggle with cross-sensor generalization, because depth sensors have different accuracy and noise properties. For some sensors we have access to large datasets with semantic annotations (ScanNet~\cite{dai2017scannet}), whilst for others, e.g. RealSense~\cite{keselman2017intel}, this is not the case. 

In this work we overcome some of the aforementioned drawbacks of 2D-3D networks-based semantic SLAM systems, see Fig.~\ref{fig:teaser}. Firstly, we propose Latent Prior Networks (LPNs) - 2D convolutional networks that leverage prior knowledge about the scene using a latent feature re-projection mechanism based on differentiable rendering, eliminating the image processing redundancy. Compared to prior methods~\cite{xiang2017darnn, ma2017multi} that re-project late features or segmentation labels, we use early features and rely on differentiable rendering. This leads to  improved quality of image-level semantic labels compared to the state-of-the-art 2D-3D networks-based method SVCNN~\cite{huang2021supervoxel}. 
 
Secondly, we propose quasi-planar over-segmentation (QPOS) for lightweight 3D map post-processing. Together with a novel SegConvNet for map post-processing this helps our system to improve semantic map annotation quality over SVCNN, matching the accuracy of 3D network-based approaches.

As a third contribution we report a significant improvement in cross-sensor generalization of a trained model based on a 2D LPN compared to a baseline 3D CNN. This may be an additional justification 2D-3D networks-based designs in semantic SLAM systems. We demonstrate our novel Semantic Mapping approach on our new RealSense dataset. See the code and data at \url{https://github.com/slamcore/semlaps}.

\begin{figure*}[t]
\vspace{0.5em}
\includegraphics[width=\linewidth]{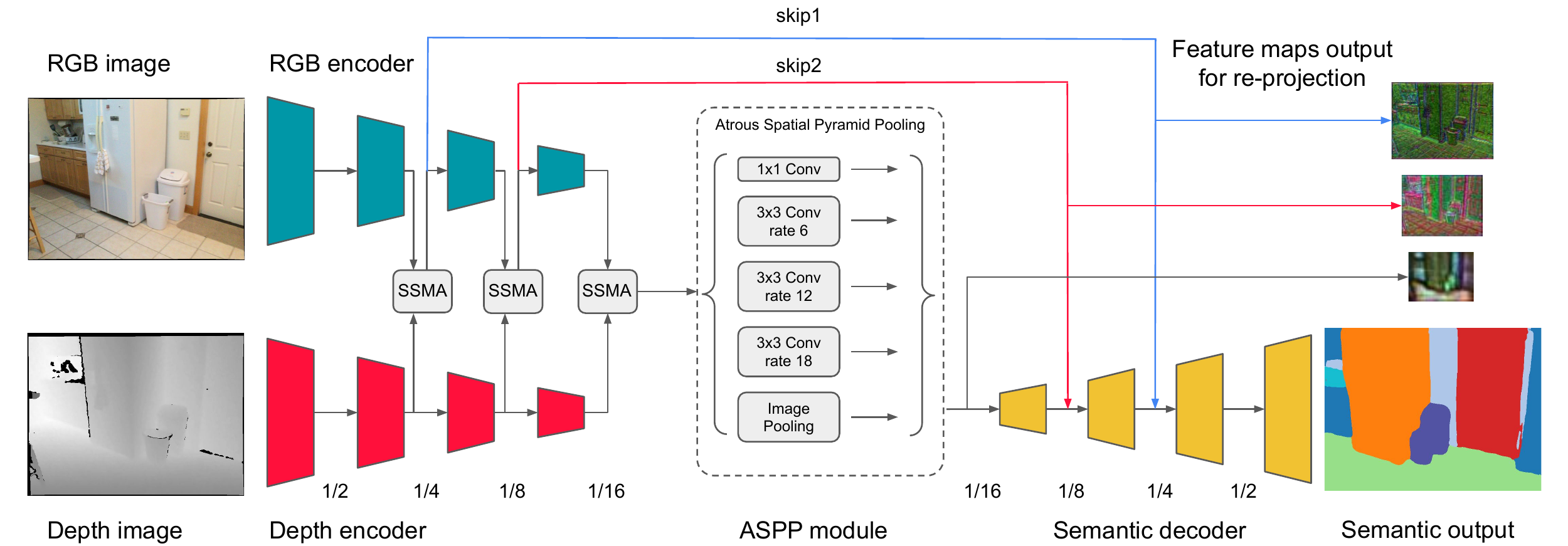}
\vspace{-4mm}
\caption{Our Latent Prior Network (LPN) architecture. For an individual frame, LPN takes in RGB and depth as input and produces the pixel-wise semantic class probabilities as well as intermediate feature maps at $\times4$, $\times8$ and $\times16$ stride. Features extracted from RGB and depth are fused at multiple resolutions with SSMA module~\cite{valada2020self}. The Atrous Spatial Pyramid Pooling (ASPP) is applied at the bottleneck followed by a lightweight decoder. 
\vspace{-6mm}
}\label{fig:lpn}
\end{figure*}
\section{Related Work}
\highlight{

{\bf 2D-3D networks-based semantic mapping}. 
Single frame RGB semantic segmentations can be lifted to 3D and fused using Bayesian~\cite{mccormac2017semanticfusion,cavallari2016semanticfusion} or geometric~\cite{narita2019panopticfusion} fusion. For frame-level fusion DA-RNN~\cite{xiang2017darnn} and MVCNet~\cite{ma2017multi} rely on depth-based re-projection of labels or late features, while our Latent Prior Network uses differentiable rendering of early features.

3D semantic label refinement is done with a CRF in ProgressiveFusion~\cite{pham2019real}, a lightweight PointConv~\cite{wu2019pointconv}-based method in FP-Conv~\cite{zhang2020fusion}  and SVCNN~\cite{huang2021supervoxel}. SVCNN defines state-of-the-art for the 2D-3D networks-based semantic mapping systems.The design of SeMLaPS is similar to SVCNN, with several major differences in 2D and 3D networks and map over-segmentation.

{\bf 3D over-segmentation.} 
Early methods for voxel grid over-segmentation were designed by extending image over-segmentation methods~\cite{moore2008superpixel,zhou2015multiscale,felzenszwalb2004efficient}. VCCS~\cite{papon2013voxel} was one of the first methods directly aimed at 3D over-segmentation, proposing a new error metric for this task. Online 3D mapping requires incremental and real-time methods. Tateno \textit{et al.}~\cite{tateno2015real} uses a 3D map as a set of object-like point clusters found using surface normals, ProgressiveFusion~\cite{pham2019real} uses a simplistic incremental voxel clustering method, while SVCNN~\cite{huang2021supervoxel} relies on a modified supervoxel segmentation~\cite{lin2018toward}.
Our proposed incremental QPOS improves over supervoxel segmentation~\cite{huang2021supervoxel,lin2018toward}.

{\bf 3D networks for semantic mapping.}
Methods of this class process a 3D reconstruction of a scene directly and produce semantic labels as output. PointNet-based semantic segmentation methods process unordered point clouds at multiple scales~\cite{qi2017pointnet}. PointConv~\cite{wu2019pointconv} and KPConv~\cite{thomas2019kpconv} propose convolution operations on point clouds. 
Sparse sub-manifold convolutions~\cite{graham20183d} and MinkowskiNet~\cite{choy20194d} process only occupied surface voxels in dense voxel grids, with reasonable memory requirements. 
BP-Net~\cite{hu2021bidirectional} leverages both a 3D network and a 2D network, linked by a feature projection mechanism. 
INS-Conv~\cite{liu2022ins} shows a way to run 3D network-based inference in an online fashion, matching the accuracy of the offline 3D networks. However, it does not produce image-level semantic labels required for other semantic SLAM tasks.
}

\section{Real-time Semantic Mapping System}
\subsection{System Overview}\label{sec:overview}
Our pipeline relies on a SLAM system, see Sec.~\ref{sec:occmap}; we process the 2D images with a novel 2D Latent Prior Network (LPN), Sec.~\ref{sec:lpn}. We over-segment the map with the novel Quasi-Planar Over-Segmentation (QPOS) described in Sec.~\ref{sec:qpos}. The Segment-Convolutional Network (SegConvNet), see Sec.~\ref{sec:scn}, gives the final output.Fig.~\ref{fig:intro} shows the overall pipeline.




\subsection{Real-Time Dense 3D Occupancy Mapping}\label{sec:occmap}
We rely on a feature-based visual inertial SLAM system (similar to \cite{Campos2021TOR}) which additionally outputs globally corrected trajectories, when loop-closures are detected and optimised. We fuse depth images and 6 Degrees of Freedom~(DoF) poses into a sub-map based 3D occupancy map. Inside each submap, occupancy information is stored in an adaptive resolution octree following \cite{funk2021multi}.

\subsection{Latent Prior Network}\label{sec:lpn}

Our LPN relies on the knowledge obtained from several frames, see Fig.~\ref{fig:lpn}. As opposed to~\cite{xiang2017darnn,ma2017multi}, we aim to inject this prior knowledge  as early as possible. 

We adopt SSMA~\cite{chen2017rethinking}, using separate encoders for RGB and depth inputs, but replace the ResNet-50 encoder with a lightweight MobileNetV3~\cite{Howard2019mobilenetv3}, keeping computational complexity in mind, see Fig.~\ref{fig:lpn}.

To enforce the prior from previous views, we propose to re-project feature maps at different resolutions to a common reference view using the depth maps and the camera poses provided by the SLAM system. We use $\times4$, $\times8$ and $\times16$ downsampled feature maps for re-projection, see Fig.~\ref{fig:lpn}. To get smooth gradient propagation, we employ  Pytorch3D for feature re-projection, see Sec. 3.3 in~\cite{ravi2020pytorch3d}. We fuse re-projected and reference view feature maps using average pooling. 

During training, at each iteration we sample $N$ adjacent frames and randomly select one as the reference view. \highlight{We randomly select the reference view instead of using the last one directly to simulate more diverse motion patterns, which serves as a data augmentation process as well.} Latent priors from $N-1$ neighbouring frames are warped to the reference view. We then compute the cross-entropy losses: $\mathcal{L}_{sem}$ from the reference view output, and $\mathcal{L}_{aux}$  neighboring views, to encourage plausible single view prediction. Our final loss is the weighted sum $\mathcal{L} = \mathcal{L}_{sem} + w\mathcal{L}_{aux}$. \highlight{We empirically found the auxiliary loss to improve around 0.7 2D mIoU points for our default setting.}

\begin{figure}
\centering
\includegraphics[width=0.9\linewidth]{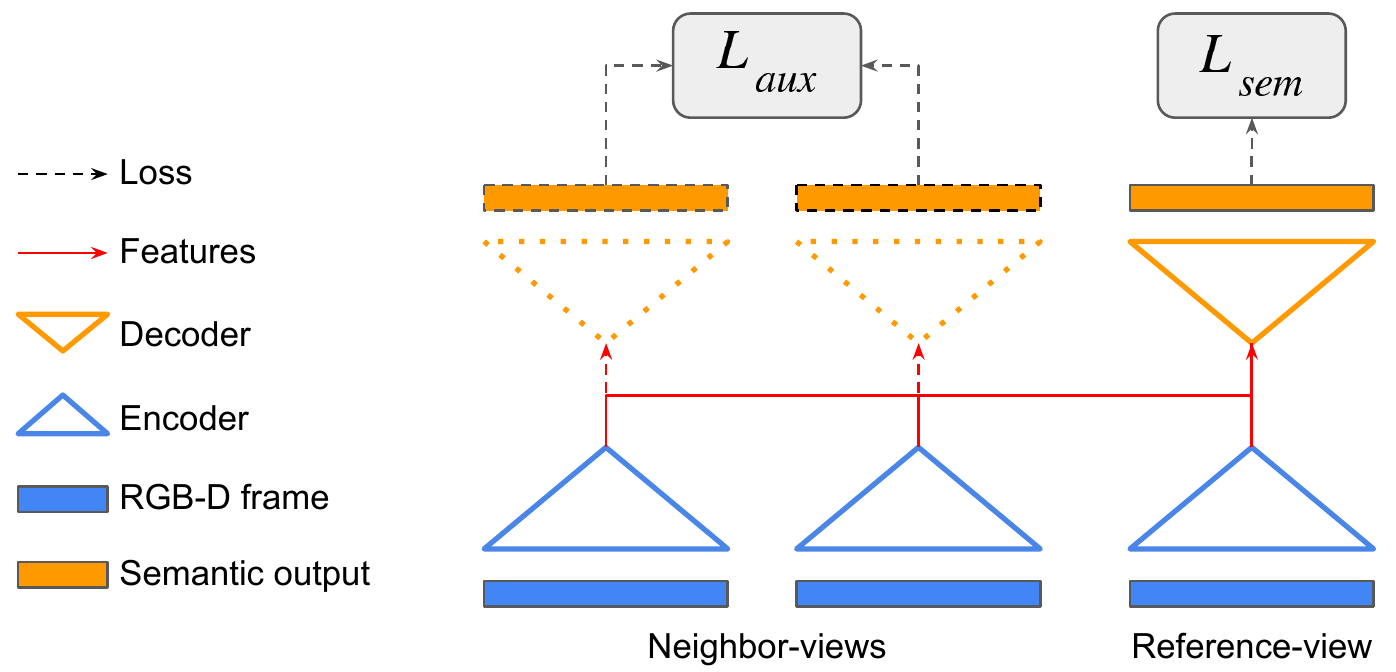}
\vspace{-0.5em}
\caption{Latent prior re-projection process. Feature maps at multiple resolutions are re-projected and fused to the reference view. At training time the views come in mini-batches. We predict semantic outputs for computing the main $\mathcal{L}_{sem}$ and auxiliary  $\mathcal{L}_{aux}$ losses. At test-time we process frames sequentially, predicting the output for the reference view only. \vspace{-16pt}}\label{fig:reprojection_mode}
\end{figure}

LPN design is flexible: firstly, although $N$ is fixed during training, at test time our LPN can take an arbitrary number of views;
secondly, 
we can do inference in sequential mode processing each frame once and reusing feature maps computed for previous view; finally, while LPN requires depth for feature re-projection and cross-view fusion, at the feature extraction stage we can rely on RGB-only input and drop the depth encoder (red part in Fig.~\ref{fig:lpn}). 




\subsection{Quasi-Planar Over-Segmentation}\label{sec:qpos}



Following~\cite{huang2021supervoxel,tateno2015real,karpathy2013object}, we aim to reduce the map cardinality by grouping voxels into segments. We extract a list of surface voxels and estimate their normals using the distance field, build an undirected weighted graph $(V, E, w)$ with voxels as vertices $V$, edges $E$ between voxels sharing a face, and weights as in~\cite{karpathy2013object}.
An over-segmentation $S$ maps voxels to segment labels. We set $S(i) = i$, sort the edges by weight as in~\cite{felzenszwalb2004efficient} and merge one by one if the cost $F(S)$ decreases:
\begin{equation}\label{eq:sizecost}
    F(S) = \frac{1}{| L(S) | } \sum_{l \in L(S)} | \sum_{i} 1_{S(i) = l} - {\bar s} |, 
\end{equation}
where $L(S)$ is the set of segment labels and ${\bar s}$ is the expected spatial segment size.

To improve the quality of segment boundaries, we apply a fast post-processing step,
assigning a voxel to a segment with the lowest association cost:
\begin{equation}\label{eq:segmentreprojection}
    C(i, l) = (p_i - p_l)^T \Sigma_l^{-1} (p_i - n_l) + \nu (1 - | n_i \cdot n_l |) ,
\end{equation}
where $p_i, n_i$ are the position and normal of the voxel $i$, $p_l, \Sigma_l, n_l$ are the center, covariance and normal of the segment $l$, and $\nu$ is a weighting factor. For efficiency, we use a KD-tree of the segment centres. 

The QPOS method has $O(|E|)$ complexity, and we use it in an incremental fashion after the map update, using only updated voxels in $V$ and $E$, trying to associate new voxels with observed segments using~(\ref{eq:segmentreprojection}) before creating new segments.  We use the QPOS results as input to the SegConvNet described next.

\subsection{Segment-Convolutional Network}\label{sec:scn}
Optimal segment sizes may differ depending on scene properties. We propose a convolution operation tailored for non-uniformly-sized segments while  SVCNN~\cite{huang2021supervoxel} assumes size uniformity.
As in PointConv~\cite{wu2019pointconv}, we define the weight-net $W(\cdot)$ as a multi-layer perceptron (MLP) 
for predicting the convolution kernel and the feature-net $\Phi(\cdot)$ as a MLP for feature processing:
\begin{equation}\label{eq:segconv}
    SF_{out}(s_x) = \sum_{s_y \in N(s_x)} W(s_y - s_x, \phi_{xy}) \Phi(SF_{in}(s_y)),
\end{equation}
where $N(s_x)$ are the neighbouring segments for $x$, $s_x, s_y$ are the segments' centers, the input segment feature $SF_{in}$ is the averaged feature within a segment, and $\phi_{xy}=[n_x \cdot n_y, r \cdot n_x,  r \cdot n_y,  n_y \cdot v, n_y \cdot w, 
r_x^y \cdot n_x, r_x^y \cdot (n_y \times n_x), \|r_x^y\|, \sigma_x, \sigma_y]$
%
%
where $n_x$ is the surface normal of segment $s_x$, $r_x^y = s_y - s_x$ is the displacement vector between the two segments, and $r=\frac{r_x^y}{\|r_x^y\|},v,w$ is the ortho-normal basis constructed by the Gram-Schmidt process from $r_x^y$ and $n_x$, $\sigma_x = \sqrt{r^T\Sigma_x r}$, $\sigma_y = \sqrt{r^T\Sigma_y r}$,
%
%
and $\Sigma_x, \Sigma_y$ represents the spatial covariances of $s_x$ and $s_y$. Our $\phi_{x,y}$ augments the view-invariant (VI) features proposed in~\cite{li2023improving} with directional variances $\sigma_x,\sigma_y$. As for the input segment features $SF_{in}$, we use averaged voxel-wise predicted class probabilities, as well as a 9-D geometric feature consisting of RGB color, position and surface normal. For each segment-convolutional layer, we use a 2-layer MLP with a hidden dimension of 8 as $W(\cdot)$ and a 2-layer MLP with a hidden dimension of 64 as $\Phi(\cdot)$. We stack 3 segment-convolutional layers to form our Segment-Convolutional Network (SegConvNet).

\begin{table*}[ht]
\vspace{1.0em}
\caption{3D Semantic segmentation on ScanNet v2 (Val). \methodname outperforms other state-of-the-art 2D-3D network-based methods. \vspace{-1.0em}}
\label{tab:scannet3d}
\begin{center}
\setlength\tabcolsep{1.5pt}
\begin{tabular}{|c|c|c|c|c|c|c|c|c|c|c|c|c|c|c|c|c|c|c|c|c|c|c|}
\hline
Method & Type & mIoU & bath & bed & bkshf & cab & chair & cntr & curt & desk & door & floor & other & pic & fridge & shwr & sink & sofa &tbl & toilet & wall & win\\
\hline
\hline
\multicolumn{23}{|c|}{\bf 3D networks} \\
\hline
MkNet~\cite{choy20194d} & offline & 72.2 & 85.9 & 81.0 & \textbf{80.7} & 66.0 & \textbf{91.1} & 63.4 & \textbf{75.9} & \textbf{64.2} & 61.4 & 94.8 & \textbf{60.0} & 29.6 & \textbf{62.0} & 69.1 & 66.8 & 81.7 & \textbf{76.3} & 92.1 & 83.3 & 59.1 \\
SparseConvNet~\cite{graham20183d} & offline & 69.3 & 86.5 & 78.6 & 76.7 & 61.2 & 89.0 & 60.9 & 70.8 & 59.1 & 60.1 & 94.5 & 47.6 & 31.5 & 47.4 & 68.9 & 62.3 & 83.3 & 68.3 & 85.4 & 82.3 & 60.0 \\
INS-CONV-m32~\cite{liu2022ins} & online & 71.5 & 84.3 & 79.3 & 78.7 & 64.4 & 90.1 & 63.6 & 72.5 & 63.5 & 60.4 & \textbf{95.2} & 55.8 & 33.8 & 52.3 & \textbf{74.5} & 63.3 & 83.2 & 74.5 & 92.8 & 84.0 & 63.3 \\
INS-CONV-m64~\cite{liu2022ins} & online & \textbf{72.4}  & \textbf{87.4} & \textbf{81.2} & 79.6 & \textbf{67.7} & 91.0 & \textbf{64.5} & 74.9 & 60.8 & \textbf{62.1} & 95.1 & 57.8 & \textbf{36.0} & 52.0 & 72.2 & \textbf{67.0} & \textbf{83.3} & 72.3 & \textbf{93.3} & \textbf{85.1} & \textbf{65.2} \\
\hline
\multicolumn{23}{|c|}{\bf 2D-3D networks} \\
\hline
PsF~\cite{pham2019real} & online & 55.0 & 65.6 & 61.2 & 65.7 & 48.6 & 68.4 & 41.7 & 54.9 & 48.9 & 47.5 & 87.1 & 43.7 & 25.7 & 41.8 & 34.5 & 53.4 & 59.8 & 54.0 & 78.9 & 70.6 & 47.0 \\
FPC~\cite{zhang2020fusion} & online & 67.2 & 85.4 & \textbf{82.3} & 64.0 & 60.9 & 75.1 & 56.0 & 64.8 & 58.2 & \textbf{64.8} & 91.9 & 46.4 & \textbf{40.6} & \textbf{64.2} & 51.7 & \textbf{63.5} & \textbf{77.9} & 68.9 & 87.0 & 83.8 & 56.3 \\
SVCNN~\cite{huang2021supervoxel} & online & 68.3 &73.4 & 78.5 & 79.1 & 60.5 & 80.6 & 59.3 & 70.4 & 59.9 & 60.5 & 91.1 & 57.8 & 35.0 & 57.5 & 75.2 & 61.3 & 72.6 & 64.4 & 86.4 & 80.5 & 61.7 \\
\methodname (ours) & online & \textbf{72.2} & \textbf{87.7} & 80.4 & \textbf{82.2} & \textbf{63.0} & \textbf{87.3} & \textbf{66.2} & \textbf{76.3} & \textbf{65.8} & 63.5 & \textbf{94.6} & \textbf{52.9} & 36.0 & 63.7 & \textbf{75.2} & 63.0 & 74.7 & \textbf{72.0} & \textbf{91.9} & \textbf{84.0} & \textbf{64.2} \\
\hline
\end{tabular} \vspace{-1.5em}
\end{center}
\end{table*}
\subsection{Semantic Mapping with RealSense}\label{sec:smr}
\highlight{ While SceneNN~\cite{hua2016scenenn} and ScanNet~\cite{dai2017scannet}, are captured with Kinect-like sensors, depth sensing accuracy of popular RealSense devices is significantly lower.} To understand the cross-sensor generalization abilites of different semantic mapping methods, we propose a four-sequence RGB-D test dataset captured using RealSense D455~\cite{keselman2017intel}. It has ground truth poses obtained using our visual inertial SLAM system and meshes reconstructed using our dense mapping system based on TSDF fusion with a voxel size of 0.01 m. Meshes are manually annotated with semantic labels in consistency with ScanNet~\cite{dai2017scannet}. There are four indoor scenes (meeting room, lab, kitchen and office lounge). We will release the dataset upon paper acceptance.
\subsection{System Implementation Details}
Finally, we provide implementation details of our entire system and the networks proposed to perform online real-time semantic mapping. 

\noindent \textbf{Latent Prior Network.} 
We train LPN on 1201 training sequences of the ScanNet v2 dataset with a step of 20 between adjacent frames, using Adam~\cite{kingma2014adam} for 20 epochs with an initial learning rate of $1e-4$ and one-cycle learning rate scheduler; We use $N=3$ and perform random scale, crop, flip, Gaussian blur and random view-order permutation for data augmentation. Training takes around 3 days on a single nVidia RTX-3090ti GPU with a batch size of 8.

\noindent \textbf{Segment-Convolutional Network.} We train the SegConvNet described in sec.~\ref{sec:scn} on the 1201 meshes of ScanNet v2 training split. We first run our QPOS for meshes with segment size ${\bar s} = 60$ vertices and transfer the vertex-wise GT labels to segment-level GT labels via majority vote. We run our trained LPN sequentially on each scene, transfer the 2D label predictions to 3D meshes using Bayesian Fusion~\cite{mccormac2017semanticfusion}. 

The above data generation process creates around 2M segments in total. We train our SegConvNet with Adam~\cite{kingma2014adam} optimizer with an initial learning rate of $5e-4$ and one-cycle scheduler. The training takes around 2 hours on a single Nvidia RTX-3090ti GPU with a batch size of 12 for 100 epochs.

\noindent \textbf{System Design Details.} We run LPN sequentially and perform Bayesian fusion at each key-frame when the SLAM system updates the map geometry. So, each frame is processed by the LPN only once, reference frames getting re-projected feature maps from non-reference frames as inputs. Next, we perform QPOS only on regions that are affected by the map update, 
updating the properties (features, segment centers, connecting topology, etc.) of the segments. Those updated segments and their K-nearest neighbors are fed to SegConvNet to predict the updated class labels.

\begin{figure*}[t]
\includegraphics[width=\linewidth]{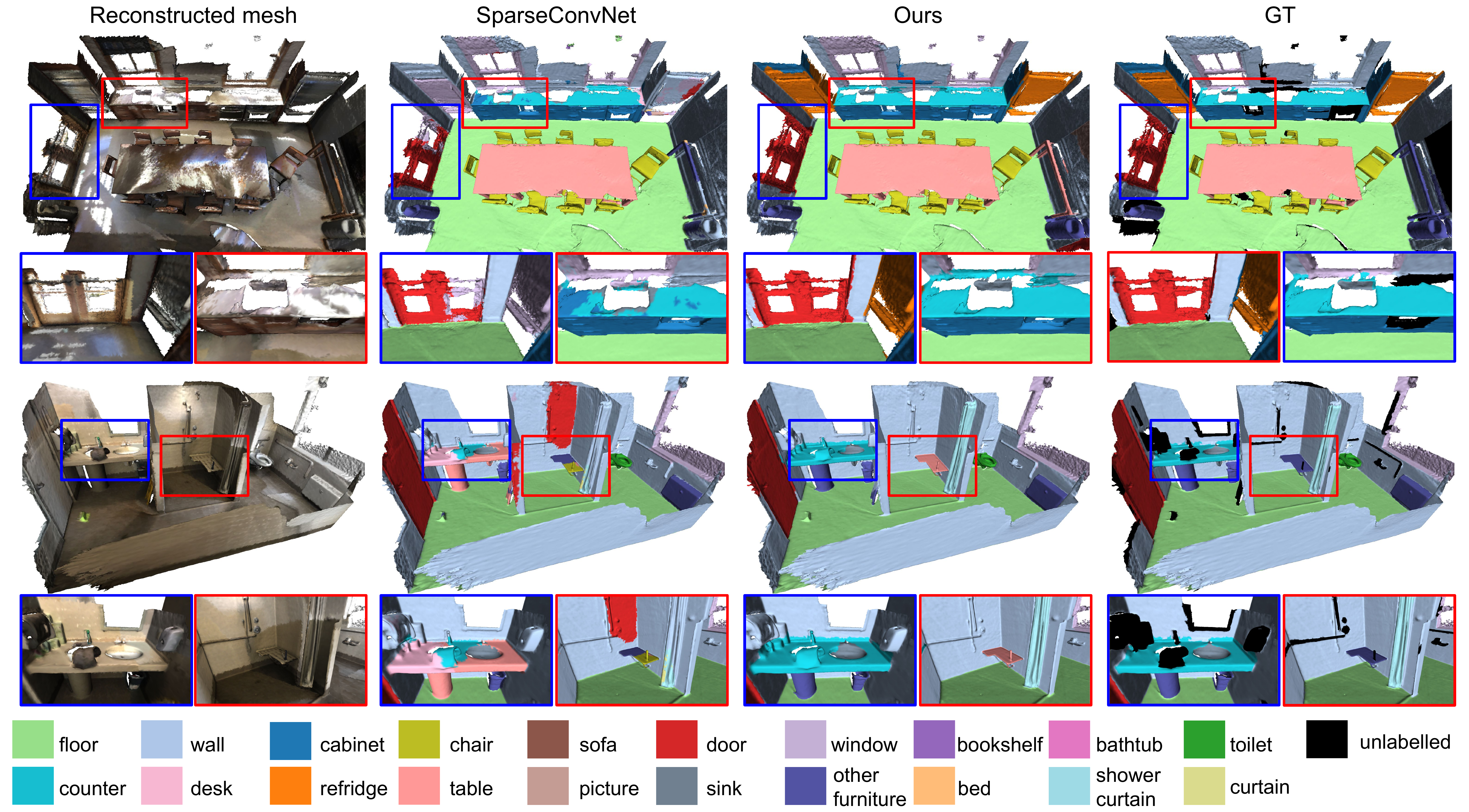}
\vspace{-4mm}
\caption{3D semantic segmentation results on ScanNet. \methodname achieves better results in classes reliant on visual features, such as refrigerator and counter (see zoomed-ins), and  produces smoother and less-fragmented  boundaries thanks to our QPOS and SegConvNet. \vspace{-1.4em}}\label{fig:scannet_comparison_3d}
\end{figure*}
\section{Experiments}
We start by introducing baselines and metrics, followed by a description of the evaluation of 2D and 3D semantics and over-segmentation, and conclude with an ablation study and timing analysis.

We use three different datasets. ScanNet~\cite{dai2017scannet} has 1201 and 312 indoor  RGB-D 
 sequences in the training and validation sub-sets, poses and meshes are obtained with BundleFusion~\cite{dai2017bundlefusion}.
SceneNN~\cite{hua2016scenenn} has 76 indoor RGB-D sequences, poses and meshes obtained using ElasticReconstruction~\cite{choi2015robust}. ScanNet and SceneNN are captured with Kinect-like sensors. Our new Semantic Mapping dataset, captured with RealSense (SMR), has 4 RGB-D sequences described in~\ref{sec:smr}. All datasets have semantically labelled meshes, and ScanNet has some frames with 2D semantic labels as well.

For 3D semantic segmentation experiments (Sec~\ref{sec:3d}), we include the following baselines: 3D network-based methods  SparseConvNet~\cite{graham20183d}, MinkowskiNet~\cite{choy20194d} and INS-CONV~\cite{liu2022ins}, which is an incremental version of SparseConvNet; 2D-3D networks-based methods   SVCNN~\cite{huang2021supervoxel}, Fusion-aware PointConv (FP-Conv)~\cite{zhang2020fusion} and ProgressiveFusion (PsF)~\cite{pham2019real}. 
For 2D semantic segmentation evaluation we use AdapNet++~\cite{valada2020self}, FuseNet~\cite{hazirbas16fusenet}, SSMA~\cite{valada2020self} and PS-res-excite~\cite{sodano2022robust}, and our closest baseline MVCNet~\cite{ma2017multi} based on our LPN backbone, in single-view inference (MVCNet-S) and in X-view inference (MVCNet-MX) \highlight{which we re-implemented}. LPN-base is trained in a single-view mode, while LPN-S and LPN-MX are trained using $N=3$ views and use one and X views for inference respectively\footnote{\highlight{Results are from respective papers apart from re-implemented MVCNet and re-trained SparseConvNet evaluated on SMR.}}.




\subsection{3D Semantic segmentation}\label{sec:3d}
\begin{figure*}[t]
\includegraphics[width=\linewidth]{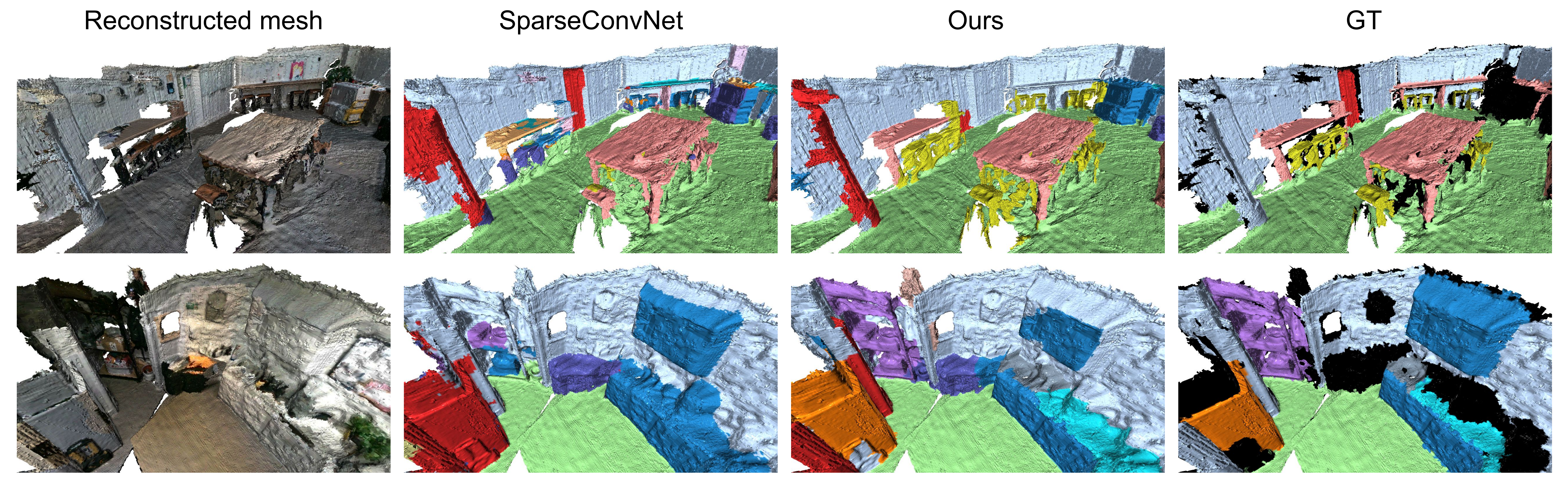}
\vspace{-1.5em}
\caption{3D semantic segmentation  on Semantic Mapping with Realsense (SMR) sequences. Our approach is more consistent especially for less frequent classes, demonstrating better cross-sensor generalization than the 3D network.\vspace{-1.6em}
}\label{fig:slamcore_comparison_3d}
\end{figure*}
We show quantitative and qualitative 3D semantic segmentation results. As mentioned before, our LPN can have multiple variants, such as using a different number of views at inference time, using depth encoder or not, mini-batch or sequential inference mode. By  default, we use LPN-M3 with both RGB and depth encoders in sequential mode for 2D inference. Our SegConvNet is trained with data generated by the same trained LPN-M3 model. If not otherwise stated, we refer to our default method as \methodname.

\noindent \textbf{ScanNet dataset.}  \methodname shows the best results among state-of-the-art 2D-3D networks-based semantic mapping methods, demonstrating big gains in segmenting classes like bathtub, chair or counter, see Tab.~\ref{tab:scannet3d}. Our proposed approach is on par with off-line and online incremental 3D networks.

Fig.~\ref{fig:scannet_comparison_3d} shows some qualitative results. From the figure it can be seen that our 2D-3D pipeline performs better in classes that rely more on visual features, such as refrigerator (first row) and counter (last row). Moreover, compared with a pure 3D-based method, our method produces smoother boundaries and fewer floating artefacts thanks to the post-processing of QPOS and SegConvNet.

\begin{table}[tp]
\caption{Online 3D Semantic segmentation on SceneNN. \vspace{-0.5em}}
\label{tab:scenenn}
\centering
\setlength\tabcolsep{1.5pt}
\begin{tabular}{C{3.5cm}|C{1.5cm}|C{1.5cm}}
\hline
Method & Type & mAcc \\
\hline
FP-Conv~\cite{zhang2020fusion} & 2D-3D & 71.5 \\
SVCNN~\cite{huang2021supervoxel} & 2D-3D & 76.9 \\
INS-CONV~\cite{liu2022ins} & 3D & 79.5   \\
Ours & 2D-3D & \textbf{80.0} \\
\hline
\end{tabular}
\vspace{-2.0em}
\end{table}

\noindent \textbf{SceneNN dataset.} We follow the protocol of SVCNN~\cite{huang2021supervoxel} and INS-CONV~\cite{liu2022ins}, using \methodname trained on ScanNet to test the generalization ability. We report average mean accuracy (mAcc $\%$) on the 48 scenes used in SVCNN. As shown in Tab.~\ref{tab:scenenn}, \methodname outperforms all the previous online semantic segmentation methods.

\noindent \textbf{SMR dataset.} We compare the SparseConvNet 3D network and the proposed \methodname-RGB (still using LPN-M3 but dropping the depth encoder), both trained on ScanNet, see Tab.~\ref{tab:slamcore}. \methodname-RGB shows significantly better cross-sensor generalization: our mIoU is better by 69\%, and mAcc is better by 6\%. Our method is better at dealing with less frequent classes, such as refrigerator, counter, chair and window, while performing on par with our baseline on frequent classes. 
Fig.~\ref{fig:slamcore_comparison_3d} shows some qualitative results. The reconstruction quality in SMR is much worse than in ScanNet or SceneNN. Apart from the classes already mentioned like refrigerator and counter, our \methodname-RGB also performs much better on thin structures and small objects, like the stools in the first row. It suggests that 2D-3D networks deal with inaccurate depth better than 3D networks.

\begin{table*}[t]
\vspace{1.0em}
\caption{3D Semantic segmentation results on Semantic Mapping with RealSense (SMR) dataset. }
\vspace{-1.0em}
\label{tab:slamcore}
\begin{center}
\setlength\tabcolsep{1.5pt}
\begin{tabular}{c|c|c|c|c|c|c|c|c|c|c|c|c|c|c|c|c|c}
\hline
Method & mAcc & mIoU & bkshf & cab & chair & cntr & desk & door & floor & other & pic & fridge & sink & sofa &tbl & wall & win \\
\hline
SparseConvNet~\cite{graham20183d} & 71.4 & 26.5 & 22.19 & 18.15 & 37.88 & 0.00 & 2.75 & 34.15 & \textbf{91.91} & 0.00 & 0.00 & 0.00 & 0.00 & \textbf{82.89} & 40.81 & 60.33 & 6.43 \\
\methodname-RGB & {\bf 75.3} & {\bf 44.7} & {\bf 30.92} & {\bf 41.13} & {\bf 53.69} & {\bf 81.62} & 0.09 & {\bf 41.70} & 87.73 & 0.00 & 0.00 & {\bf 82.86} & {\bf 73.70} & 40.29 & {\bf 54.08} & {\bf 62.98} &  {\bf 20.01} \\
\hline
\end{tabular}
\end{center}
\vspace{-2.0em}
\end{table*}

\vspace{-0.5em}
\subsection{2D Semantic segmentation}

In this section we evaluate 2D networks on the ScanNet v2 dataset, see Tab.~\ref{tab:semantic}. 
In the single-view RGB-D inference mode, LPN-base has the lowest performance; the MVCNet~\cite{ma2017multi} has better metrics, while our proposed LPN-S achieves the best accuracy. In the multi-view inference mode, we see that our LPN-M3 outperforms MVCNet-M3, both using three views, while increasing the number of views up to 7 improves the LPN results further. Comparing with the other architectures, we see that our LPN backbone has reasonable performance, while relying on computationally efficient MobileNetV3, when other approaches use VGG-16 or ResNet50. 
LPN-M7 outperforms all single-view baselines. 
\vspace{-1.5em}
\begin{table}[t]
  \centering
  \caption{Performance of different 2D semantic segmentation models on ScanNet v2 (Val). 
  Best result in bold. \vspace{-0.5em}} \label{tab:scannet2D}
  \begin{tabular}{C{3.0cm}C{2.0cm}C{1.5cm}c}
    \toprule
    Method   & Backbone     & Test views      & mIoU \\
    \midrule
    \multicolumn{4}{c}{\bf Single-view trained} \\
    \midrule
    AdapNet++~\cite{valada2020self}             & ResNet-50           & 1 & 54.6 \\
    FuseNet~\cite{hazirbas16fusenet}             & VGG-16           & 1 & 56.7 \\
    SSMA~\cite{valada2020self}                  & ResNet-50           & 1 & 66.1 \\
    PS-res-excite~\cite{sodano2022robust}         & ResNet-50           & 1 & \textbf{69.8} \\
    LPN-base & MobileNetV3 & 1 & 65.1 \\
    \midrule
    \multicolumn{4}{c}{\bf Multi-view trained} \\
    \midrule
    LPN-S & MobileNetV3 & 1 & 66.5 \\
    MVCNet-S & MobileNetV3 & 1 & 66.2 \\
    LPN-M3 & MobileNetV3 & 3 & 69.0 \\
    MVCNet-M3 & MobileNetV3 & 3 & 68.6 \\
    LPN-M5 & MobileNetV3 & 5 & 69.7 \\
    LPN-M7 & MobileNetV3 & 7 & \textbf{70.1} \\
    \bottomrule
  \end{tabular}

  \label{tab:semantic}
  \vspace{-0.5em}
\end{table}

\subsection{Over-Segmentation}
We run the proposed QPOS and  Supervoxel Clustering~\cite{huang2021supervoxel,lin2018toward}, on ScanNet val.  To compute the over-segmentation error~(OSE) we find  semantic labels for segments (supervoxels) from ground truth vertex labels using majority vote, propagate  labels back to vertices, and compute mean accuracy. The under-segmentation error~(USE)~\cite{papon2013voxel} is a standard error metric.
As shown in Tab.~\ref{tab:oseval}, our  QPOS outperforms Supervoxel Clustering by 6.2\% in USE and by $2.2\%$ in OSE, producing a similar average number of segments. 

\begin{table}
\caption{3D Over-Segmentation Evaluation on ScanNet v2 (Val). \vspace{-0.5em}}
\label{tab:oseval}
\begin{center}
\setlength\tabcolsep{1.5pt}
\begin{tabular}{c|c|c|c|c|c}
\hline
Method &
\begin{tabular}[x]{@{}c@{}}
US Err.\\
\%
\end{tabular} & 
\begin{tabular}[x]{@{}c@{}}
OS Err.\\
\%
\end{tabular} & 
\begin{tabular}[x]{@{}c@{}}
Avg\\
Segments
\end{tabular}  & 
\begin{tabular}[x]{@{}c@{}}
Avg\\
Seg. Size
\end{tabular} 
&
\begin{tabular}[x]{@{}c@{}}
Std\\
Seg. Size
\end{tabular}
\\
\hline
Supervoxel~\cite{lin2018toward} & 37.8 & 5.6 & 1760 & 90 & 39 \\
QPOS  & {\bf 31.6} & {\bf 3.4}  & 1760 & 90 & 30 \\
\hline
\end{tabular}
\end{center}
\vspace{-2.0em}
\end{table}


\begin{table}[t]
  \caption{\highlight{Ablation study on the effect of different modules on ScanNet v2 (Val). We show mIoU ($\%$) after each stage.} \vspace{-0.2em}}
  \centering
    \highlight{
    \begin{tabular}{cccc}
    \toprule
    \multirow{2}{*}{2D Net} & \multirow{2}{*}{OS} & BF & Final \\
    & & mIoU $\%$ & mIoU $\%$ \\
    \midrule
    LPN-M3 & QPOS & \textbf{64.5} & 
    \textbf{72.2} \\
    LPN-base& QPOS & 63.5 
    & 70.7 \\
    SSMA~\cite{valada2020self} & QPOS & 59.1 & 68.5 \\
    LPN-M3 & SV~\cite{lin2018toward}& \textbf{64.5} 
    & 70.2 \\
    LPN-base & SV~\cite{lin2018toward}& 63.5 
    & 69.0 \\
    SSMA~\cite{valada2020self} & SV~\cite{lin2018toward} & 59.1 & 67.0 \\
    \bottomrule
  \end{tabular}
  }
  \label{tab:ablation_modules}
  \vspace{-0.5em}
\end{table}
\begin{table}[t]
  \caption{Ablation study on the effect of SegConvNet. All networks are trained with data generated with our default LPN-M3. \vspace{-0.5em}}
  \centering
  \begin{tabular}{C{5.5cm}C{1.5cm}}
    \toprule
    Model & mIoU $\%$ \\
    \midrule
    SegConvNet w/o VI-feature & 71.0 \\
    SegConvNet w/o dir. variance & 71.5 \\
    SegConvNet w/o geom. feature & 70.1 \\
    SegConvNet FULL (K=64) & \textbf{72.2} \\
    \midrule
    SegConvNet K=16 & 71.2 \\
    SegConvNet K=32 & 71.8 \\
    SegConvNet K=96 & 72.1 \\
    \bottomrule
  \end{tabular}

  \label{tab:ablation_segconv}
  \vspace{-0.5em}
\end{table}
\vspace{-0.5em}
\highlight{
\subsection{Ablation study}

We conduct ablation studies to investigate the effectiveness of different components in \methodname.

\noindent\textbf{Effect of LPN and QPOS.} We first study the effect of Latent Prior Network (LPN) and Quasi-Planar Over-segmentation (QPOS), see Tab.~\ref{tab:ablation_modules}. We show the mIoU after Bayesian fusion as \textbf{BF mIoU} and the mIoU of the final output as \textbf{Final mIoU}. We start by replacing our default LPN with other 2D network variants. The first row shows that QPOS+SegConvNet improve the results by 7.7 mIoU\%, justifying the importance of our proposed 3D post-processing. Our default LPN-M3 is significantly better compared with LPN-base (second row), which indicates the effectiveness of multi-view training and incorporating latent priors.In the third row, the final mIoU drops further to 68.5 \% after replacing our LPN-base with SSMA~\cite{valada2020self}. In all cases, the final result is significantly better than the Bayesian fusion, highlighting the universal applicability of the proposed QPOS+SegConvNet step. Next, in rows 4-6 we replace the proposed QPOS method with Supervoxel clustering~\cite{lin2018toward}. QPOS  consistently yields better final mIoU results compared to the baseline. 
}


\begin{table}[h!]
\caption{Runtime (ms) of \textbf{(a)} LPN-M3; \textbf{(b)} whole \methodname. \vspace{-0.5em}}
\label{tab:timing}
\begin{center}
\setlength\tabcolsep{1.5pt}
\begin{tabular}{c|c}
\hline
\multicolumn{2}{c}{\textbf{(a) LPN-M3 run-time breakdown}} \\
Stage & Time (ms) \\
\hline
Encoder & 7.0 \\
Feature re-proj. & 11.8 \\
Decoder & 1.7 \\
Overall & 20.5 \\
\hline
\end{tabular}
\begin{tabular}{c|c}
\hline
\multicolumn{2}{c}{\textbf{(b) \methodname system run-time}} \\
Stage & Time (ms) \\
\hline
LPN & 20.5 \\
Bayesian fusion & 0.4 \\
QPOS & 11.6 \\
SegConvNet & 4.6 \\
\hline
\end{tabular}
\end{center}
\vspace{-2.2em}
\end{table}
\begin{figure}[h!]
\centering
\includegraphics[width=0.85\linewidth]{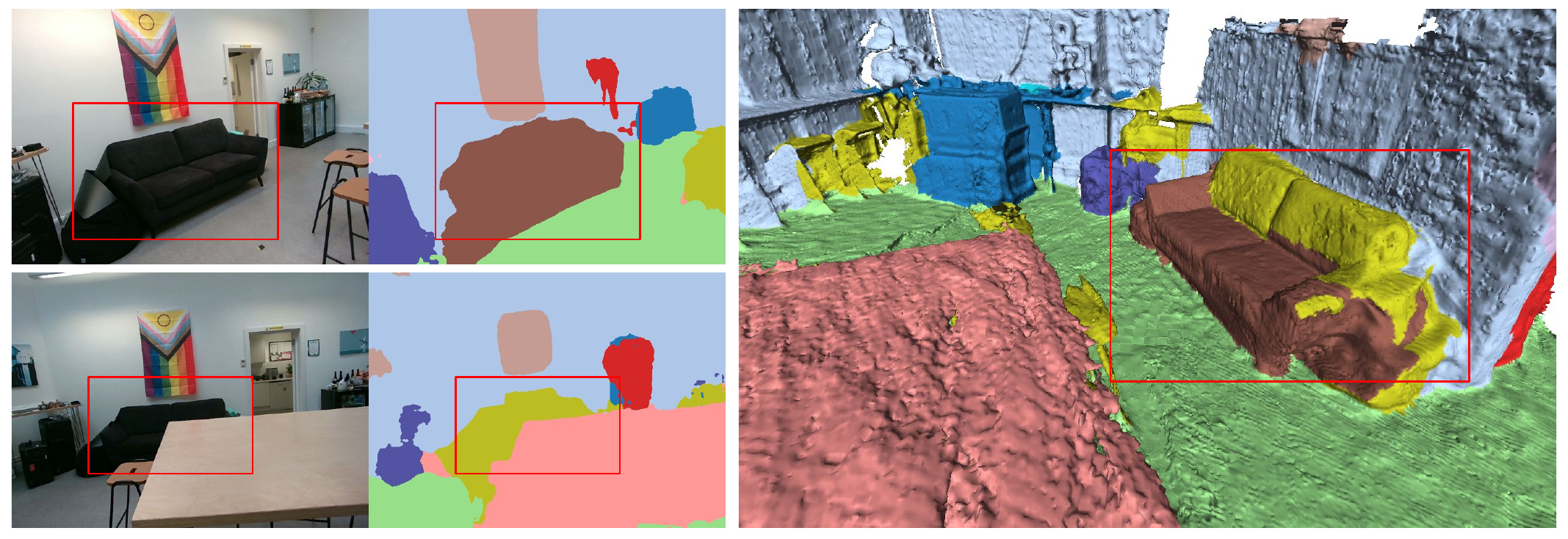}
\vspace{-0.2em}
\caption{An example failure case of \methodname. \textbf{Top Left:} A good view for LPN, \textbf{Bottom-left:} A bad view for LPN. \textbf{Right:} Final 3D results. Heavy occlusion causes LPN output wrong labels, which SegConvNet is unable to correct. \vspace{-1.5em}}\label{fig:failure_case}
\end{figure}

\noindent\textbf{Effect of SegConvNet.}  We first analyze the importance of input features to the weight-net $W(\cdot)$ and input segment feature. We train our SegConvNet with data generated by our default LPN-M3 variant, but removing $\phi_{xy}$ 
from the weight-net input and the 9-D geometric features from the input segment feature. As shown in Tab.~\ref{tab:ablation_segconv}, both VI-features and the directional variance are important for the weight-net to predict better convolution weights. Also it can be seen that 2D-only features are not descriptive enough to achieve the best result (70.1 vs. 72.2).
%
We also studied the effect of $K$. It can be seen that increasing $K$ from 16 to 64 constantly improves the result, but the improvement saturates when further increasing $K$ to 96. 

\highlight{
\noindent\textbf{Failure Case.} A major limitation of \methodname is that it heavily relies on the 2D semantic output, which could easily fail under difficult situations such as heavy occlusion and extreme motion, see Fig.~\ref{fig:failure_case}. In  sequence shown in Fig.~\ref{fig:failure_case}, the black sofa (red box) is severely occluded by the table in 920 out of 1420 frames in which it is observed, which caused the LPN to wrongly label it as chair (yellow) in those frames. Although one of the motivations of LPN is to mitigate this kind of situation by leveraging multi-view consistency, this sequence proves too difficult to handle. We believe that adopting better view selection and temporal fusion strategy should help solve this issue.}

\subsection{System Timing}

To evaluate the run-time performance of \methodname, following~\cite{narita2019panopticfusion, liu2022ins} we profile the timing for each stage of our method on \texttt{scene0645\_01}, a typical large scene in ScanNetv2. All experiments are conducted on a desktop computer with Intel Core-i7 12700K CPU and a Nvidia RTX-3090ti GPU. As \methodname consists of 4 stages: LPN, Bayesian fusion, QPOS and SegConvNet, we report the average timing for each of the 4 stages in ~\ref{tab:timing}. We also report the run-time breakdown for our LPN.

As shown in Tab.~\ref{tab:timing} (a), our default LPN-M3 model could run at 50~Hz. Thanks to our lightweight encoder backbone and sequential operating mode, the feature extraction (encoder forward pass) and decoder forward pass only take 7~ms and 1.7~ms respectively. The feature re-projection stage presents as a major bottleneck which requires 11.8~ms for processing feature maps from the other two neighboring views, but it could be optimized by replacing PyTorch3D with  a better implementation. Tab.~\ref{tab:timing} (b) shows timing for all stages.  \methodname achieves 25~Hz performance, which is more than enough for online real-time semantic mapping with our visual inertial SLAM system.

\section{Conclusion}

We presented \methodname, a real-time online semantic mapping system that follows a 2D-3D pipeline. It benefits from fusing latent features from historical views using a novel Latent Prior Network (LPN), while Quasi-Planar-Over-Segmentation (QPOS) and Segment-Convolutional Network (SegConvNet) improve the final results to the same level as 3D offline methods  while still maintaining real-time performance. \methodname achieves better cross-sensor generalization than 3D-only networks.

\highlight{Currently, \methodname cannot handle environments with frequent changes or dynamic elements. Future work could leverage the 2D-3D design to skip 2D dynamic features in the re-projection and global 3D map. Alternatively optical or scene flow field could be used for differentiable re-projection.}


\bibliographystyle{IEEEtran}
\bibliography{IEEEexample}

\begin{thebibliography}{10}
\providecommand{\url}[1]{#1}
\csname url@rmstyle\endcsname
\providecommand{\newblock}{\relax}
\providecommand{\bibinfo}[2]{#2}
\providecommand\BIBentrySTDinterwordspacing{\spaceskip=0pt\relax}
\providecommand\BIBentryALTinterwordstretchfactor{4}
\providecommand\BIBentryALTinterwordspacing{\spaceskip=\fontdimen2\font plus
\BIBentryALTinterwordstretchfactor\fontdimen3\font minus
  \fontdimen4\font\relax}
\providecommand\BIBforeignlanguage[2]{{%
\expandafter\ifx\csname l@#1\endcsname\relax
\typeout{** WARNING: IEEEtran.bst: No hyphenation pattern has been}%
\typeout{** loaded for the language `#1'. Using the pattern for}%
\typeout{** the default language instead.}%
\else
\language=\csname l@#1\endcsname
\fi
#2}}

\bibitem{mccormac2017semanticfusion}
J.~McCormac, A.~Handa, A.~Davison, and S.~Leutenegger, ``{SemanticFusion}:
  Dense {3D} semantic mapping with convolutional neural networks,'' in
  \emph{IEEE Intl. Conf. on Robotics and Automation (ICRA)}, 2017, pp.
  4628--4635.

\bibitem{ren2022visual}
Y.~Ren, B.~Xu, C.~L. Choi, and S.~Leutenegger, ``Visual-inertial multi-instance
  dynamic slam with object-level relocalisation,'' in \emph{IEEE/RSJ Intl.
  Conf. on Intelligent Robots and Systems (IROS)}.\hskip 1em plus 0.5em minus
  0.4em\relax IEEE, 2022, pp. 11\,055--11\,062.

\bibitem{tian2022kimera}
Y.~Tian, Y.~Chang, F.~H. Arias, C.~Nieto-Granda, J.~P. How, and L.~Carlone,
  ``Kimera-multi: Robust, distributed, dense metric-semantic slam for
  multi-robot systems,'' \emph{{IEEE} Trans. Robotics}, vol.~38, no.~4, 2022.

\bibitem{kantaros2022perception}
Y.~Kantaros, S.~Kalluraya, Q.~Jin, and G.~J. Pappas, ``Perception-based
  temporal logic planning in uncertain semantic maps,'' \emph{{IEEE} Trans.
  Robotics}, vol.~38, no.~4, pp. 2536--2556, 2022.

\bibitem{narita2019panopticfusion}
G.~Narita, T.~Seno, T.~Ishikawa, and Y.~Kaji, ``Panopticfusion: Online
  volumetric semantic mapping at the level of stuff and things,'' in
  \emph{IEEE/RSJ Intl. Conf. on Intelligent Robots and Systems (IROS)}.\hskip
  1em plus 0.5em minus 0.4em\relax IEEE, 2019, pp. 4205--4212.

\bibitem{huang2021supervoxel}
S.-S. Huang, Z.-Y. Ma, T.-J. Mu, H.~Fu, and S.-M. Hu, ``Supervoxel convolution
  for online 3d semantic segmentation,'' \emph{ACM Trans. Graph.}, vol.~40,
  no.~3, pp. 1--15, 2021.

\bibitem{zhang2020fusion}
J.~Zhang, C.~Zhu, L.~Zheng, and K.~Xu, ``Fusion-aware point convolution for
  online semantic 3d scene segmentation,'' in \emph{IEEE Conf. on Computer
  Vision and Pattern Recognition (CVPR)}, 2020, pp. 4534--4543.

\bibitem{grinvald2019volumetric}
M.~Grinvald, F.~Furrer, T.~Novkovic, J.~J. Chung, C.~Cadena, R.~Siegwart, and
  J.~Nieto, ``Volumetric instance-aware semantic mapping and {3D} object
  discovery,'' \emph{(IEEE) Robotics and Automation Letters}, vol.~4, no.~3,
  pp. 3037--3044, 2019.

\bibitem{xiang2017darnn}
Y.~Xiang and D.~Fox, ``{DA-RNN}: Semantic mapping with data associated
  recurrent neural networks,'' in \emph{Robotics: Science and Systems (RSS)},
  2017.

\bibitem{ma2017multi}
L.~Ma, J.~St{\"u}ckler, C.~Kerl, and D.~Cremers, ``Multi-view deep learning for
  consistent semantic mapping with {RGB-D} cameras,'' in \emph{IEEE/RSJ Intl.
  Conf. on Intelligent Robots and Systems (IROS)}.\hskip 1em plus 0.5em minus
  0.4em\relax IEEE, 2017, pp. 598--605.

\bibitem{han2020occuseg}
L.~Han, T.~Zheng, L.~Xu, and L.~Fang, ``Occuseg: Occupancy-aware 3d instance
  segmentation,'' in \emph{IEEE Conf. on Computer Vision and Pattern
  Recognition (CVPR)}, 2020, pp. 2940--2949.

\bibitem{choy20194d}
C.~Choy, J.~Gwak, and S.~Savarese, ``4d spatio-temporal convnets: Minkowski
  convolutional neural networks,'' in \emph{IEEE Conf. on Computer Vision and
  Pattern Recognition (CVPR)}, 2019, pp. 3075--3084.

\bibitem{graham20183d}
B.~Graham, M.~Engelcke, and L.~Van Der~Maaten, ``3d semantic segmentation with
  submanifold sparse convolutional networks,'' in \emph{IEEE Conf. on Computer
  Vision and Pattern Recognition (CVPR)}, 2018, pp. 9224--9232.

\bibitem{bescos2021dynaslam}
B.~Bescos, C.~Campos, J.~D. Tard{\'o}s, and J.~Neira, ``Dynaslam ii:
  Tightly-coupled multi-object tracking and slam,'' \emph{(IEEE) Robotics and
  Automation Letters}, vol.~6, no.~3, pp. 5191--5198, 2021.

\bibitem{liu2022ins}
L.~Liu, T.~Zheng, Y.-J. Lin, K.~Ni, and L.~Fang, ``{INS-Conv}: Incremental
  sparse convolution for online 3d segmentation,'' in \emph{IEEE Conf. on
  Computer Vision and Pattern Recognition (CVPR)}, 2022, pp. 18\,975--18\,984.

\bibitem{dai2017scannet}
A.~Dai, A.~X. Chang, M.~Savva, M.~Halber, T.~Funkhouser, and M.~Nie{\ss}ner,
  ``{ScanNet}: Richly-annotated 3d reconstructions of indoor scenes,'' in
  \emph{IEEE Conf. on Computer Vision and Pattern Recognition (CVPR)}, 2017,
  pp. 5828--5839.

\bibitem{keselman2017intel}
L.~Keselman, J.~Iselin~Woodfill, A.~Grunnet-Jepsen, and A.~Bhowmik, ``Intel
  realsense stereoscopic depth cameras,'' in \emph{IEEE Conf. on Computer
  Vision and Pattern Recognition Workshops (CVPRW)}, 2017, pp. 1--10.

\bibitem{valada2020self}
A.~Valada, R.~Mohan, and W.~Burgard, ``Self-supervised model adaptation for
  multimodal semantic segmentation,'' \emph{Intl. J. of Computer Vision}, vol.
  128, no.~5, pp. 1239--1285, 2020.

\bibitem{cavallari2016semanticfusion}
T.~Cavallari and L.~Di~Stefano, ``Semanticfusion: Joint labeling, tracking and
  mapping,'' in \emph{Eur. Conf. on Computer Vision (ECCV)}, 2016, pp.
  648--664.

\bibitem{pham2019real}
Q.-H. Pham, B.-S. Hua, T.~Nguyen, and S.-K. Yeung, ``Real-time progressive 3d
  semantic segmentation for indoor scenes,'' in \emph{IEEE Workshop on
  Applications of Computer Vision (WACV)}.\hskip 1em plus 0.5em minus
  0.4em\relax IEEE, 2019, pp. 1089--1098.

\bibitem{wu2019pointconv}
W.~Wu, Z.~Qi, and L.~Fuxin, ``{PointConv}: Deep convolutional networks on 3d
  point clouds,'' in \emph{IEEE Conf. on Computer Vision and Pattern
  Recognition (CVPR)}, 2019, pp. 9621--9630.

\bibitem{moore2008superpixel}
A.~P. Moore, S.~J. Prince, J.~Warrell, U.~Mohammed, and G.~Jones, ``Superpixel
  lattices,'' in \emph{IEEE Conf. on Computer Vision and Pattern Recognition
  (CVPR)}.\hskip 1em plus 0.5em minus 0.4em\relax IEEE, 2008, pp. 1--8.

\bibitem{zhou2015multiscale}
Y.~Zhou, L.~Ju, and S.~Wang, ``Multiscale superpixels and supervoxels based on
  hierarchical edge-weighted centroidal voronoi tessellation,'' \emph{IEEE
  Transactions on Image Processing}, vol.~24, no.~11, pp. 3834--3845, 2015.

\bibitem{felzenszwalb2004efficient}
P.~F. Felzenszwalb and D.~P. Huttenlocher, ``Efficient graph-based image
  segmentation,'' \emph{Intl. J. of Computer Vision}, vol.~59, pp. 167--181,
  2004.

\bibitem{papon2013voxel}
J.~Papon, A.~Abramov, M.~Schoeler, and F.~Worgotter, ``Voxel cloud connectivity
  segmentation-supervoxels for point clouds,'' in \emph{IEEE Conf. on Computer
  Vision and Pattern Recognition (CVPR)}, 2013, pp. 2027--2034.

\bibitem{tateno2015real}
K.~Tateno, F.~Tombari, and N.~Navab, ``Real-time and scalable incremental
  segmentation on dense slam,'' in \emph{IEEE/RSJ Intl. Conf. on Intelligent
  Robots and Systems (IROS)}.\hskip 1em plus 0.5em minus 0.4em\relax IEEE,
  2015, pp. 4465--4472.

\bibitem{lin2018toward}
Y.~Lin, C.~Wang, D.~Zhai, W.~Li, and J.~Li, ``Toward better boundary preserved
  supervoxel segmentation for 3d point clouds,'' \emph{ISPRS journal of
  photogrammetry and remote sensing}, vol. 143, pp. 39--47, 2018.

\bibitem{qi2017pointnet}
C.~R. Qi, H.~Su, K.~Mo, and L.~J. Guibas, ``Pointnet: Deep learning on point
  sets for 3d classification and segmentation,'' in \emph{Proceedings of the
  IEEE conference on computer vision and pattern recognition}, 2017, pp.
  652--660.

\bibitem{thomas2019kpconv}
H.~Thomas, C.~R. Qi, J.-E. Deschaud, B.~Marcotegui, F.~Goulette, and L.~J.
  Guibas, ``Kpconv: Flexible and deformable convolution for point clouds,'' in
  \emph{Proceedings of the IEEE/CVF international conference on computer
  vision}, 2019, pp. 6411--6420.

\bibitem{hu2021bidirectional}
W.~Hu, H.~Zhao, L.~Jiang, J.~Jia, and T.-T. Wong, ``Bidirectional projection
  network for cross dimension scene understanding,'' in \emph{Proceedings of
  the IEEE/CVF Conference on Computer Vision and Pattern Recognition}, 2021,
  pp. 14\,373--14\,382.

\bibitem{Campos2021TOR}
C.~Campos, R.~Elvira, J.~J.~G. Rodríguez, J.~M. M.~Montiel, and J.~D.~Tardós,
  ``Orb-slam3: An accurate open-source library for visual, visual–inertial,
  and multimap slam,'' \emph{{IEEE} Trans. Robotics}, vol.~37, no.~6, pp.
  1874--1890, 2021.

\bibitem{funk2021multi}
N.~Funk, J.~Tarrio, S.~Papatheodorou, M.~Popovi{\'c}, P.~F. Alcantarilla, and
  S.~Leutenegger, ``Multi-resolution 3d mapping with explicit free space
  representation for fast and accurate mobile robot motion planning,''
  \emph{(IEEE) Robotics and Automation Letters}, vol.~6, no.~2, pp. 3553--3560,
  2021.

\bibitem{chen2017rethinking}
L.-C. Chen, G.~Papandreou, F.~Schroff, and H.~Adam, ``Rethinking atrous
  convolution for semantic image segmentation,'' \emph{arXiv preprint
  arXiv:1706.05587}, 2017.

\bibitem{Howard2019mobilenetv3}
A.~Howard, M.~Sandler, G.~Chu, L.-C. Chen, B.~Chen, M.~Tan, W.~Wang, Y.~Zhu,
  R.~Pang, V.~Vasudevan, Q.~V. Le, and H.~Adam, ``Searching for mobilenetv3,''
  in \emph{Proceedings of the IEEE/CVF International Conference on Computer
  Vision (ICCV)}, October 2019.

\bibitem{ravi2020pytorch3d}
N.~Ravi, J.~Reizenstein, D.~Novotny, T.~Gordon, W.-Y. Lo, J.~Johnson, and
  G.~Gkioxari, ``Accelerating 3d deep learning with {PyTorch3D},''
  \emph{arXiv:2007.08501}, 2020.

\bibitem{karpathy2013object}
A.~Karpathy, S.~Miller, and L.~Fei-Fei, ``Object discovery in 3d scenes via
  shape analysis,'' in \emph{IEEE Intl. Conf. on Robotics and Automation
  (ICRA)}, 2013, pp. 2088--2095.

\bibitem{li2023improving}
X.~Li, W.~Wu, X.~Z. Fern, and L.~Fuxin, ``Improving the robustness of point
  convolution on k-nearest neighbor neighborhoods with a viewpoint-invariant
  coordinate transform,'' in \emph{IEEE Workshop on Applications of Computer
  Vision (WACV)}, 2023, pp. 1287--1297.

\bibitem{hua2016scenenn}
B.-S. Hua, Q.-H. Pham, D.~T. Nguyen, M.-K. Tran, L.-F. Yu, and S.-K. Yeung,
  ``Scenenn: A scene meshes dataset with annotations,'' in \emph{2016 fourth
  international conference on 3D vision (3DV)}.\hskip 1em plus 0.5em minus
  0.4em\relax Ieee, 2016, pp. 92--101.

\bibitem{kingma2014adam}
D.~P. Kingma and J.~Ba, ``Adam: A method for stochastic optimization,''
  \emph{arXiv preprint arXiv:1412.6980}, 2014.

\bibitem{dai2017bundlefusion}
A.~Dai, M.~Nie{\ss}ner, M.~Zollh{\"o}fer, S.~Izadi, and C.~Theobalt,
  ``Bundlefusion: Real-time globally consistent 3d reconstruction using
  on-the-fly surface reintegration,'' \emph{TOG}, vol.~36, no.~4, p.~1, 2017.

\bibitem{choi2015robust}
S.~Choi, Q.-Y. Zhou, and V.~Koltun, ``Robust reconstruction of indoor scenes,''
  in \emph{IEEE Conf. on Computer Vision and Pattern Recognition (CVPR)}, 2015,
  pp. 5556--5565.

\bibitem{hazirbas16fusenet}
C.~Hazirbas, L.~Ma, C.~Domokos, and D.~Cremers, ``Fusenet: Incorporating depth
  into semantic segmentation via fusion-based cnn architecture,'' in
  \emph{Asian Conference on Computer Vision (ACCV)}, November 2016.

\bibitem{sodano2022robust}
M.~Sodano, F.~Magistri, T.~Guadagnino, J.~Behley, and C.~Stachniss, ``Robust
  double-encoder network for rgb-d panoptic segmentation,'' \emph{arXiv
  preprint arXiv:2210.02834}, 2022.

\end{thebibliography}

\end{document}